\documentclass[runningheads]{llncs}

\usepackage{graphicx}
\usepackage{amsmath}
\usepackage{amssymb} 

\begin{document}

\title{Generative Model-based Simulation of Driver Behavior when Using Control Input Interface for Teleoperated Driving in Unstructured Canyon Terrains
\thanks{This work was supported by Agency for Defense Development Grant funded by the Korean Government (2023).}}

\titlerunning{Generative Model-based Simulation of Teleoperated Driver Behavior}

\author{Hyeonggeun Yun\inst{1} \and
Younggeol Cho\inst{2} \and
Jinwon Lee\inst{1} \and
Arim Ha\inst{1} \and
Jihyeok Yun\inst{1}}

\authorrunning{H. Yun et al.}

\institute{Agency for Defense Development, Daejeon, Republic of Korea\\
\email{\{yhg8423,leejw509,arimida86,jihyeok.yun\}@gmail.com}
\and
Korea Advanced Institute of Science and Technology, Daejeon, Republic of Korea\\
\email{rangewing@kaist.ac.kr}}

\maketitle

\begin{abstract}
Unmanned ground vehicles (UGVs) in unstructured environments mostly operate through teleoperation. To enable stable teleoperated driving in unstructured environments, some research has suggested driver assistance and evaluation methods that involve user studies, which can be costly and require lots of time and effort. A simulation model-based approach has been proposed to complement the user study; however, the models on teleoperated driving do not account for unstructured environments.
Our proposed solution involves simulation models of teleoperated driving for drivers that utilize a deep generative model. Initially, we build a teleoperated driving simulator to imitate unstructured environments based on previous research and collect driving data from drivers. Then, we design and implement the simulation models based on a conditional variational autoencoder (CVAE). Our evaluation results demonstrate that the proposed teleoperated driving model can generate data by simulating the driver appropriately in unstructured canyon terrains.

\end{abstract}

\section{Introduction}

Teleoperated driving of unmanned ground vehicles (UGVs) is a prominent example of teleoperation in unstructured environments. One of the difficulties in teleoperated driving is stability, which is driving while sustaining consistent states in response to abrupt changes in the local environment \cite{opiyo2021review}. In unstructured environments, ensuring stable driving is essential to reduce collision risks and increase vehicle safety. However, teleoperation is more complex than commercial driving, as drivers lack physical feedback from the vehicle regarding speed, acceleration, and posture \cite{park2021design}. Furthermore, the difficulty of teleoperated driving increases in unstructured environments since roads are not standardized, and there are diverse elements, such as canyons, obstacles, and potholes. Consequently, teleoperated driving in unstructured environments requires extensive practice and is difficult for drivers.

To support teleoperated driving, previous research proposed driver assistance methods for collision avoidance and stable driving \cite{cho2023goondae,storms2017shared}. User studies and human-in-the-loop experiments have been used to evaluate driver assistance methods. However, conducting studies with drivers requires a huge effort and time, making repeated or iterative experiments challenging. Human-in-the-loop experiments and teleoperated driving with drivers, in particular, are high-cost. Thus, an evaluation approach based on a simulation model of users is proposed to complement the user study \cite{murray2022simulation}.

Previously, modeling and simulating of drivers have been studied in terms of driving behavior. Wang et al. classified and clustered driving behaviors using a machine learning-based approach for personalized compensation for speed tracking errors \cite{wang2020driver}. Li et al. proposed a human driving model that predicts the steering behavior of drivers in the teleoperation of UGV with different speeds \cite{li2022modeling}. They used ACT-R cognitive architecture, a two-point steering model, and a far-point control model to simulate human steering behavior. Schnelle et al. also suggested a driver steering model which can adapt a driving behavior of an individual driver \cite{schnelle2016driver}. Existing research has primarily centered on the development of simulation models for driving behaviors on urban roads, such as lane changing and vehicle following, using rule-based approaches. Unfortunately, these models have a limited capacity to reflect the driving behavior of multiple drivers, and their applicability to mimicking teleoperated driving of drivers in unstructured environments is also restricted.

In this paper, we solve this problem using a self-supervised generative model. Based on previous research \cite{murray2021forward}, we propose a simulation model of teleoperated driving commands based on a conditional variational autoencoder (CVAE) in unstructured environments, particularly unstructured canyon terrains. 
First, we collected driving data from drivers using a teleoperated driving simulator, including control inputs, dynamic state information of a platform, and environmental information. The data is then used to train and validate our simulation models. Second, we design a forward simulation model and an inverse simulation model for teleoperated driving based on CVAE. CVAE is a deep conditional generative model that produces stochastic inferences based on the inputs and generates realistic output predictions \cite{sohn2015learning}. We first train the forward simulation model to generate environmental information and dynamic state information, then train the inverse simulation model to generate control inputs for drivers. 

\section{Teleoperated Driving Dataset Collection}\label{sim_dataset}

\begin{figure}[h]
  \centering
  \includegraphics[width=0.85\textwidth]{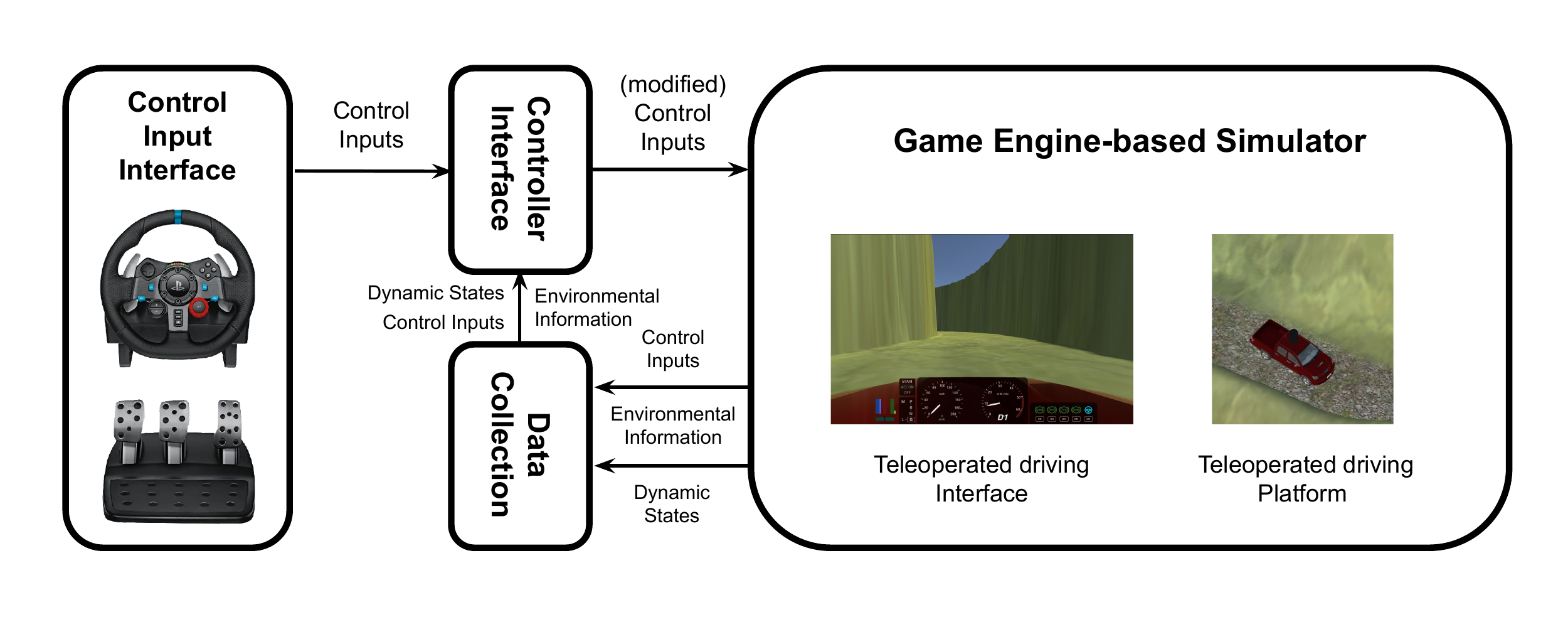}
    \caption{Game engine-based simulator configuration with input interface.}
  \label{fig:simulator}
\end{figure}

Designing a teleoperated driving model requires driving data from drivers in unstructured environments. Thus, we employed an off-road teleoperated driving simulator that could reflect real-world physical effects from previous work to simulate teleoperated driving in the unstructured environment \cite{cho2023goondae} (See Figure \ref{fig:simulator}). The simulator is implemented using the Unity Pro game engine and Vehicle Physics Pro automotive simulation engine. The simulator provided a truck-like wheeled platform with a LiDAR sensor that replicates Velodyne 16-channel 3D LiDAR and driving interfaces including a speedometer and a tachometer. Drivers used a Logitech G29 racing wheel as a control input interface.

\begin{figure}
  \centering
  \includegraphics[width=0.85\textwidth]{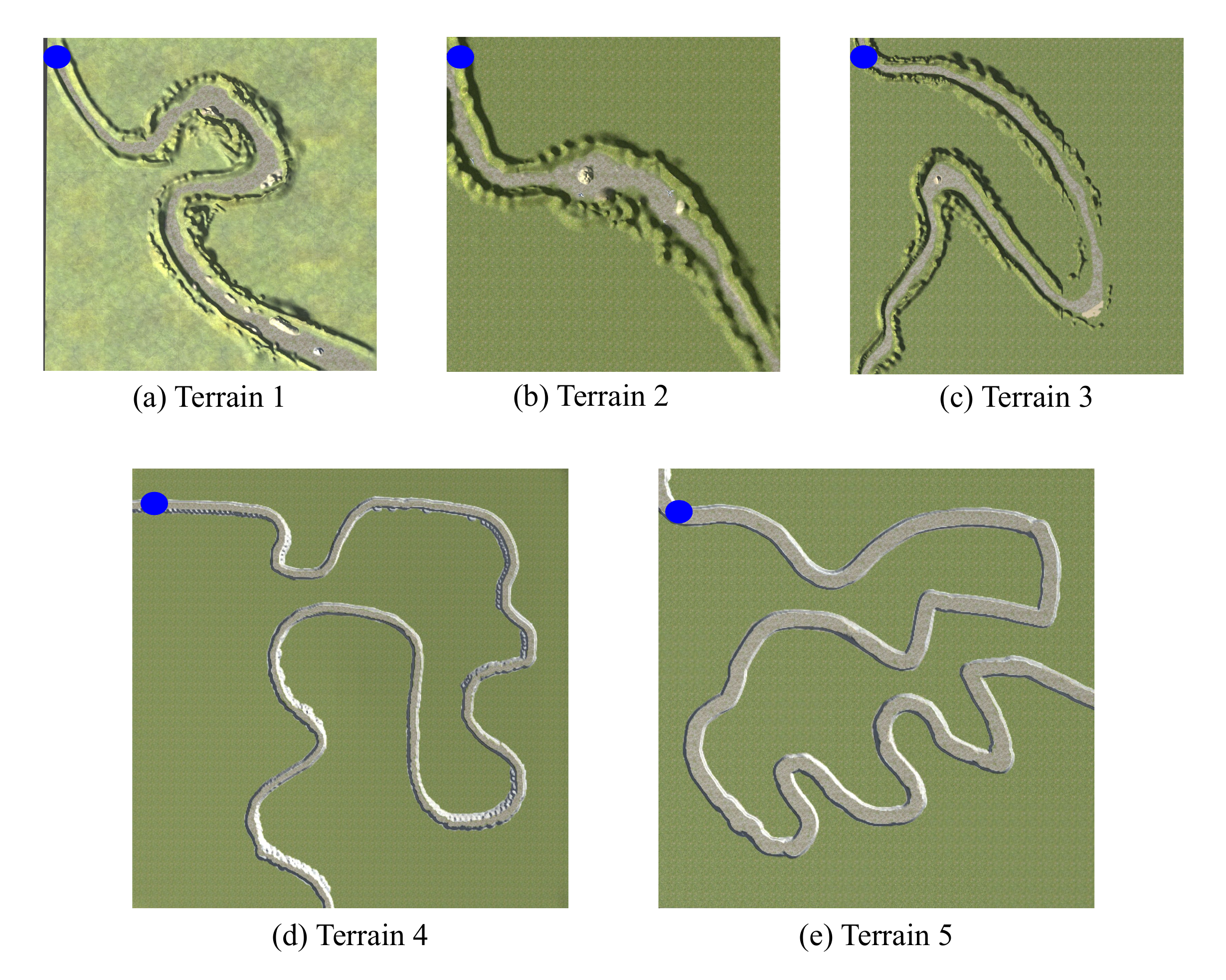}
  \caption{Five terrains (maps) include both curved and linear pathways for data collection and model evaluation. Blue circle denotes the starting point of each terrain.}
  \label{fig:terrains}
\end{figure}

To simulate the teleoperated driving, we collected the following simulator data: control inputs, dynamic state information, and environmental information. The control inputs consisted of steering data and pedal data from the controller. The dynamic state information consisted of the current speed, orientation (yaw, roll, pitch), and current position of the platform in the simulator. The environmental information in this study refers to LiDAR Point Cloud data captured by the LiDAR sensor. The simulator stores all data every 100 milliseconds because we contemplated a generation model that utilizes environmental data.

To collect driving data, we utilized five terrains that imitated canyons and rough terrains (See Figure \ref{fig:terrains}). As we concentrated on unstructured environments and canyons, the terrains' pathways were unpaved, laneless, and surrounded by steep mountains.

We collected driving data from 5 drivers  (experienced drivers) with experience using the simulator and 14 drivers  (inexperienced drivers) with no experience using the simulator. The experienced drivers continuously collected driving data from three terrains (Figure \ref{fig:terrains}.a, b, c). The inexperienced drivers operated the wheeled platform on the two terrains (Figure \ref{fig:terrains}.d, e) to collect driving data. We used a driving dataset of experienced drivers for the forward simulation model and a dataset of inexperienced drivers for the inverse simulation model.

\section{TELEOPERATED DRIVING MODEL}

We implemented a teleoperated driving model based on the driving data of drivers. Our objective is to simulate the teleoperated driving of drivers so that we can readily measure and predict the teleoperated driving performance without drivers participating in teleoperated driving experiments directly. To achieve the goal, we considered a simulation model based on a deep generative model adopting a conditional variational autoencoder (CVAE). A previous study has demonstrated that a CVAE is useful at simulating physical movements based on sensor inputs and generating various outputs from given conditions \cite{murray2021forward}. We considered that a CVAE-based model could generate control inputs for multiple drivers by utilizing the relationship between current and past data.

Therefore, we designed a teleoperated driving simulation model based on a CVAE that can mimic the teleoperated driving behavior of drivers. Since we used the training and inference framework of previous research \cite{murray2021forward}, we designed a forward simulation model and an inverse simulation model with a CVAE. The forward model uses the previous data from the nine-time steps and the control inputs from the current time step to generate the environmental data and dynamic state data for the current time step, and the inverse model uses the previous data from the nine-time steps and the environmental data and dynamic state data from the current time step to generate the control inputs for the current time step. 

\subsection{Preprocessing}

We preprocessed the collected dataset in accordance with prior research in order to enhance the effectiveness of training and quality of data \cite{cho2023goondae,choe2015obstacle,li20203d,naranjo2007cooperative}. In preprocessing, the dataset was divided into control inputs consisting of steering input and acceleration/brake pedal input, dynamic state information consisting of orientation data and speed data, and environmental information consisting of LiDAR Point Cloud data. For each $i^{th}$ time step, we created a Control Vector $c_i$, a State Vector $s_i$, and an Environment Vector $e_i$ through preprocessing. 

First, by normalizing the steering input data, acceleration pedal input data, and brake pedal input data, we generate a Control Vector $c_i$ for the control inputs \cite{naranjo2007cooperative}. The steering input data was normalized to a range of 0-1. The acceleration pedal input data was scaled to a range of 0-1, while the brake pedal input data was scaled to -1-0. They were then combined as the pedal input data, which was subsequently normalized from -1-1 to 0-1.

We create a State Vector $s_i$ for the dynamic state information of the platform by normalizing the current speed and orientation data. The orientation data was normalized to a range of 0-1 from 0-360 degrees, and the speed data was normalized to a range of 0-1 from 0-30 m/s, because the current speed of data did not exceed 30 m/s.

In the environmental information, an Environment Vector $e_i$ was constructed using LiDAR Point Cloud data. The LiDAR Point Cloud data was converted to cylindrical coordinates and obstacle points were identified by detecting points with a slope greater than the threshold (which was set to 45 degrees for this paper) along the same azimuth \cite{choe2015obstacle,li20203d}. The obstacle point data was then transformed into a vector depicting the distance between the obstacle and the platform along the forward 180 degrees. The distance values were then normalized to a range of 0-1 on the assumption that the utmost distance was 50 meters.

\subsection{Training and Inference Approach with Two Simulation Models}

As we mentioned above, we trained two simulation models: a forward simulation model and an inverse simulation model. The forward model generates observed data derived from user interactions, while the inverse model generates user interactions derived from observed data \cite{moon2023amortized,murray2021forward}. Generally, the forward model is derived from principles and theories based on psychology and physics, and the inverse model is derived from a machine learning-based model based on users' behavior data. However, in our proposed model, we also used a machine learning-based model for the forward model since we thought that the observed sensor data could be derived from the control inputs and sensor data of previous time steps.

Thus, we designed two CVAE-based simulation models. The training and inference approach with two simulation models consisted of three steps.

\begin{figure}[h]
  \centering
  \includegraphics[width=0.9\textwidth]{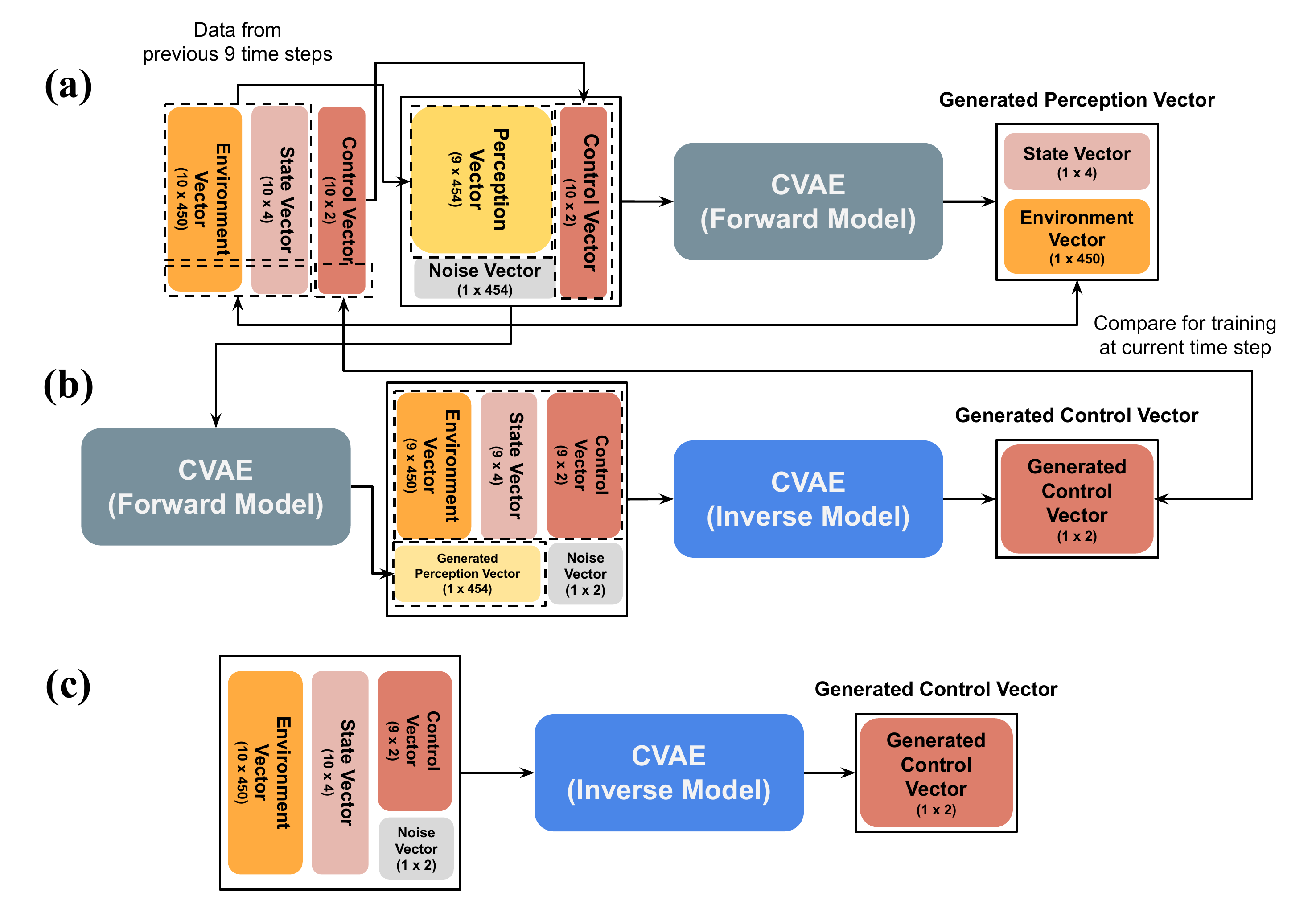}
    \caption{Training and inference approach with a forward simulation model and an inverse simulation model.}
  \label{fig:training_method}
\end{figure}

\begin{itemize}
    \item[(1)] First, we create a Perception Vector by combining an Environment Vector and State. We utilize the Perception Vector and Control Vector of the previous nine-time steps and the Control Vector of the current time step as a condition vector for the CVAE, and use a Noise Vector of the same size as the Perception Vector of the current time step as input. The forward CVAE model generates the Perception vector corresponding to the current time step. To generate accurate perception vectors, we use a dataset of 102,073 time steps generated by the experienced drivers for training (Figure \ref{fig:training_method}.a).
    \item[(2)] Second, we use the Perception Vector generated by the forward CVAE model used in step (1) and the Perception Vector and Control Vector from the previous nine-time steps as a condition vector, and a Noise Vector of the same size as the Control Vector at the current time step as input. The Control Vector at the current time step is generated by the inverse CVAE model. To generate diverse drivers' behavior, we use a dataset of 88,955 time steps generated from inexperienced drivers for training (Figure \ref{fig:training_method}.b).
    \item[(3)] Third, we utilize the inverse CVAE model trained in step (2) to generate a Control Vector for the current time step from the Perception Vectors and Control Vectors of the previous nine-time steps and the Perception Vector for the current time step. The inverse CVAE model simulates the control inputs of drivers represented by the generated control vector (Figure \ref{fig:training_method}.c).
\end{itemize}

We used time-series data from ten consecutive time steps, including the current time step and the nine-time steps preceding it.  The condition vector was constructed by combining the State Vector, Environment Vector, and Control Vector. We also used a random vector as a noise vector $\tilde{n}$, and the sample vector $z$ was set to the same size as the noise vector. The dataset was divided into a training dataset for updating the model's parameters and a validation dataset for verifying the model's performance. The model was trained for 1,000 epochs using a batch size of 2,048. The initial learning rate was set to $1 \times 10 ^ {-3}$, and it was decreased to a tenth every 300 epochs using a step decay scheduler in order to enhance training performance. Mean Squared Error (MSE) function and Kullback–Leibler (KL) divergence were used for loss functions, whereas VAE and CVAE typically use Binary Cross Entropy (BCE) function and KL divergence for loss functions. In addition, since it is crucial to restoring the vector at the current time step based on previous information, we only compared a loss between generated vector at the current time step $c_{gen, t}$ for training. The training was conducted on a GPU server equipped with the NVIDIA Tesla A100.

\subsection{Model Architecture}

\begin{figure}[h]
  \centering
  \includegraphics[width=0.95\textwidth]{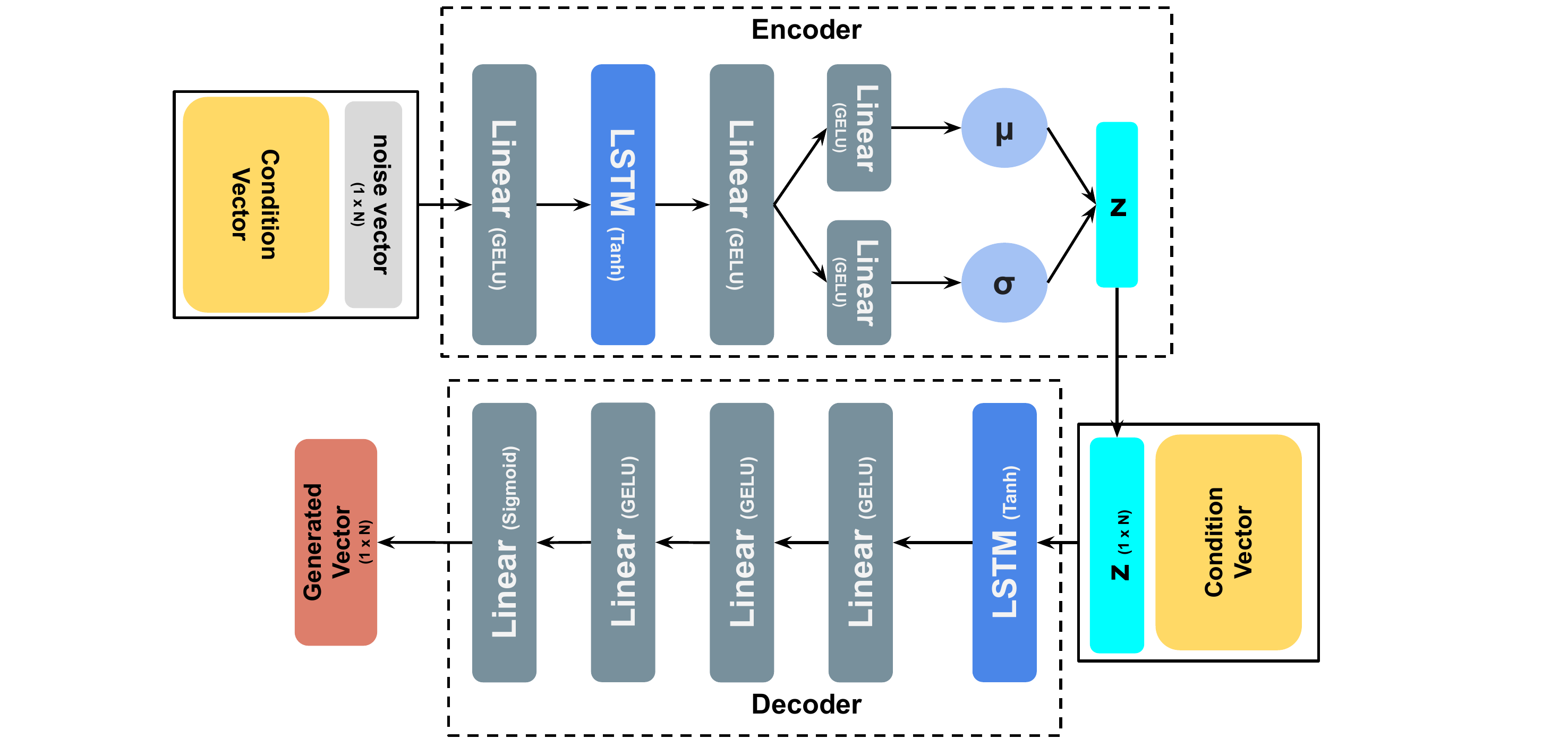}
    \caption{CVAE-based teledriving model architecture for a forward simulation model and an inverse simulation model. $N$ is the size of the generated vector to simulate.}
  \label{fig:sim_architecture}
\end{figure}

The architecture of the CVAE-based teleoperated driving model is illustrated in Figure \ref{fig:sim_architecture}. Our model comprises an encoder with four Linear layers and an LSTM layer, as well as a decoder with four Linear layers and an LSTM layer.

In the encoder, we first generate a Condition Vector $q = (q_{t-9}, q_{t-8}, ..., q_t)$ and a Noise Vector $ \tilde{n} \in \mathbb{R}^{N}$ where $t$ represents the current time step and $N$ represents the size of the generated vector to simulate. Then, the combination of Noise Vector $ \tilde{n} $ and Condition Vector $q$ generates an input vector $ x = (q_{t-9}, q_{t-8}, ..., q_t \oplus \tilde{n})$ where $\oplus$ represents concatenation. We first extract time-series features $x_{enc}$ from $x$ with two Linear layers and an LSTM layer.

\begin{equation}
    x_{enc} = Linear_{enc, 2}(LSTM_{enc}(Linear_{enc, 1}(x)))
\end{equation}

Then, two Linear layers generate a mean vector $\mu$ and a variance vector $\sigma$ that are typically employed in variational autoencoder (VAE).

\begin{equation}
    \mu = Linear_{enc, 3}(x_{enc})
\end{equation}

\begin{equation}
    \sigma = Linear_{enc, 4}(x_{enc})
\end{equation}

The model then generates a sample vector $z = (z_{t-9}, z_{t-8}, ..., z_{t})$ from the mean vector $\mu$ and variance vector $\sigma$.

\begin{equation}
    z = \mu + \sigma^2 \odot \epsilon
\end{equation}

where $\epsilon \sim N(0, 1^2)$ and $\odot$ is an element-wise product.

In the decoder, we only need $z_{t}$, so the $z_t$ vector and the Condition Vector $q$ are combined to generate the vector $x' = (q_{t-9}, q_{t-8}, ..., q_t \oplus z_t)$.

Then, sequential features $r_{dec, 1}$ are extracted from $x'$ by an LSTM layer of the decoder.

\begin{equation}
    r_{dec, 1} = LSTM_{dec}(x')
\end{equation}

Finally, four Linear layers of the decoder then generate a Vector $g_{gen} = (g_{gen, t-9}, g_{gen, t-8}, ..., g_{gen, t}) \in \mathbb{R}^{N}$. We use $g_{gen, t}$ vector to simulate state and environmental information or control inputs corresponding to the current time step.

\begin{equation}
    r_{dec, 2} = Linear_{dec, 3}(Linear_{dec, 2}(Linear_{dec, 1}(r_{dec, 1})))
\end{equation}

\begin{equation}\label{eqn:last_linear}
    g_{gen} = Linear_{dec, 4}(r_{dec, 2})
\end{equation}

The activation function for Linear layers, excluding the last Linear layer of the decoder, was the Gaussian error linear unit (GELU) function, while the activation function for LSTM layers was the Tanh function. The last Linear layer of the decoder utilized the sigmoid function as an activation function to generate a vector in the range 0-1.

\section{Model Evaluation}
\subsection{Evaluation Design}
To evaluate the capacity of our CVAE-based teledriving model to simulate the teleoperated driving behavior of drivers, we compared the driving performance of the simulated model to that that of real drivers. We collected teleoperated driving data of the simulation model from the canyon terrain (Figure \ref{fig:terrains}.e) where drivers drove. As a dataset of drivers' teleoperated driving consists of a total of 28 driving data operated by 14 drivers twice, the simulation model also drove 28 times in the same terrain. After arriving at the destination, driving data was added to the dataset for purposes of comparison.

To compare the driving performance of drivers to that of the simulation model, we employed four metrics based on previous research \cite{cho2023goondae,knappe2007driving,knapper2015comparing,neumeier2019teleoperation}: the standard deviation of lateral position (SDLP), the standard deviation of speed (SDS), the average speed (AvgSpeed), and the driving completion time (DCT). For all metrics, lower scores mean superior driving performance.
   
\subsection{Evaluation Result}

\begin{figure}[h]
  \centering
  \includegraphics[width=0.5\linewidth]{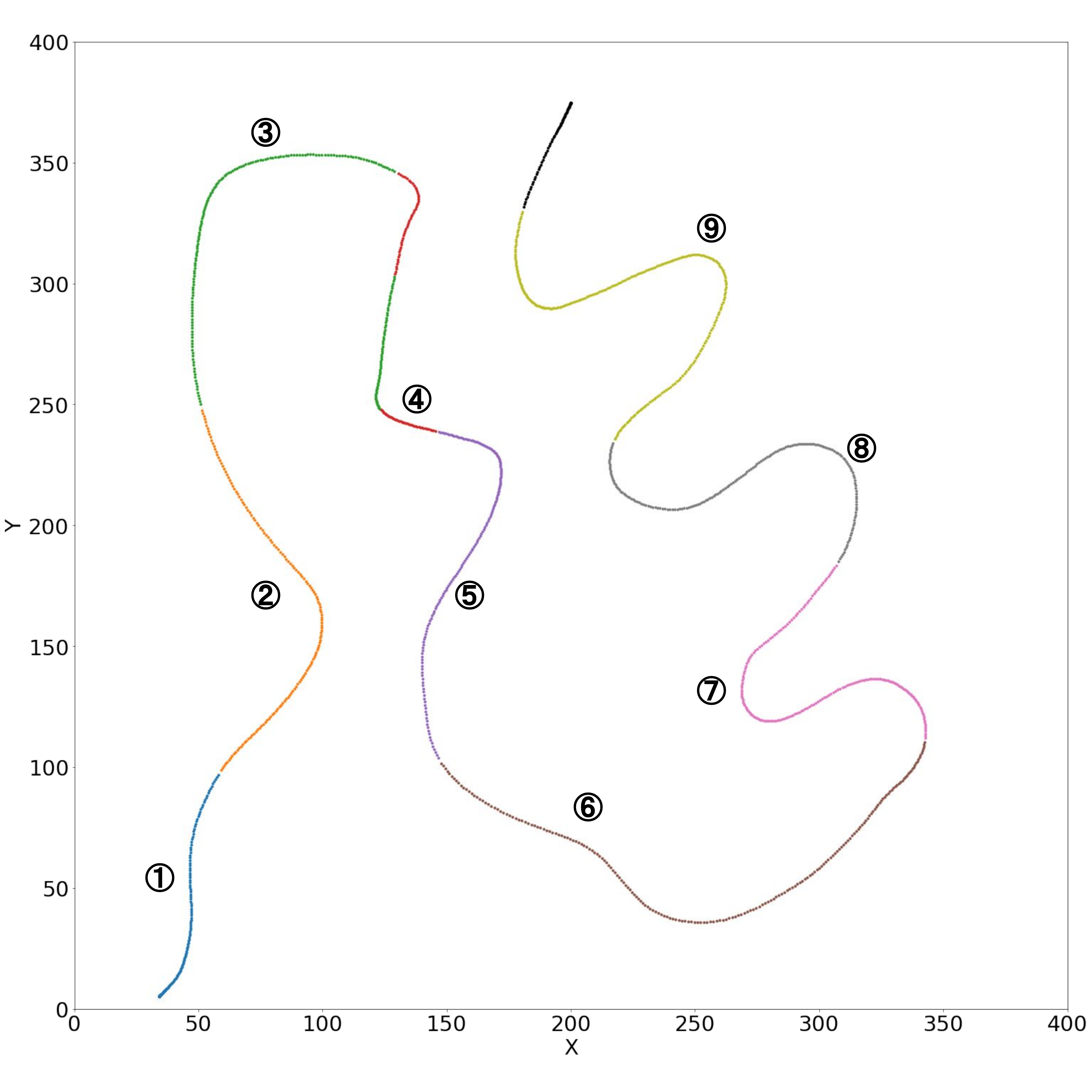}
  \caption{Terrain divisions for drivers and simulation model performance comparison}
  \label{fig:sim_segmentation}
\end{figure}

\begin{figure*}[h]
  \centering
  \includegraphics[width=0.8\linewidth]{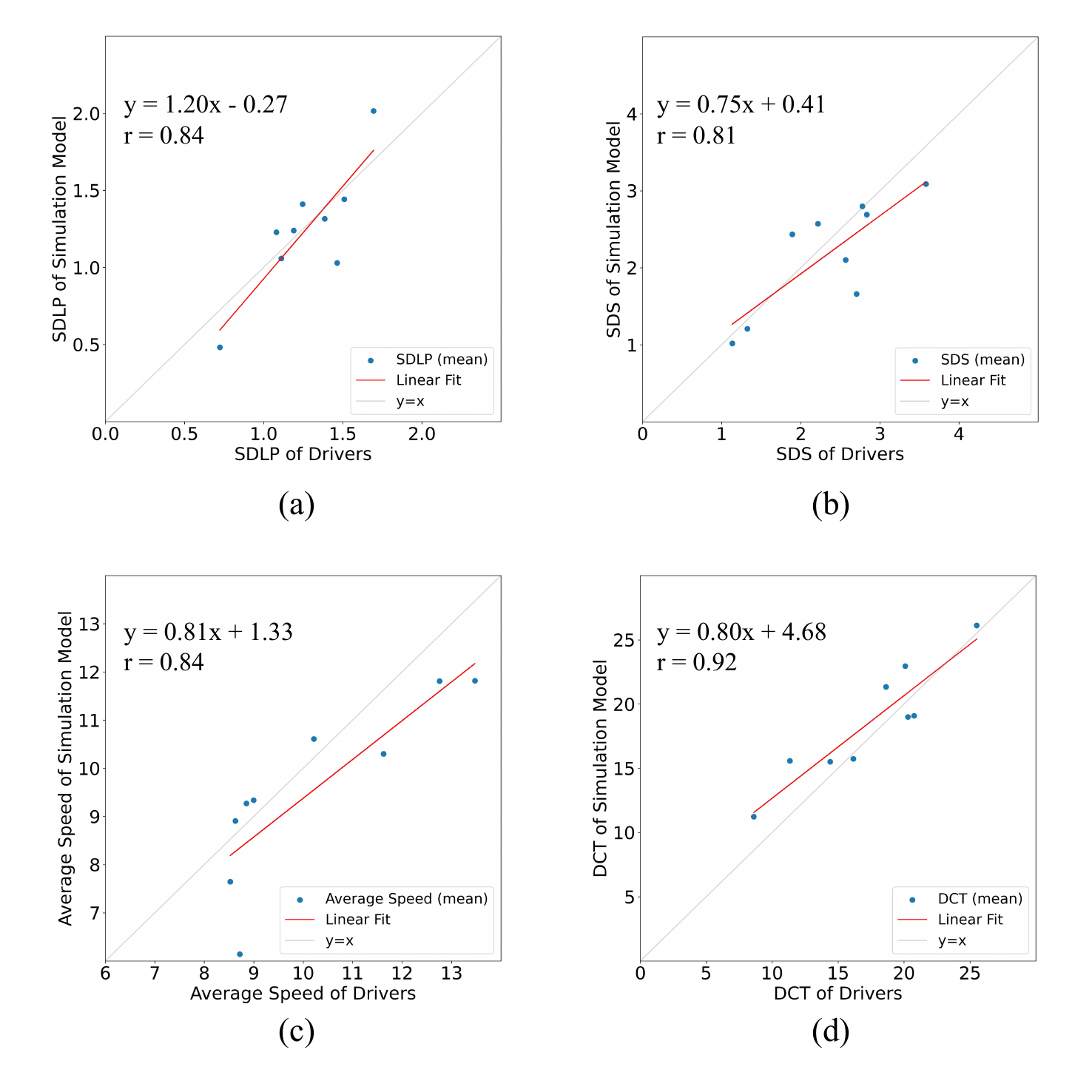}
  \caption{Nine divisions analysis of the relationship between drivers and simulation model performance. A blue dot represents two driving performances in each division: the drivers' and the simulation model's.}
  \label{fig:sim_y_x}
\end{figure*}

\begin{figure*}[h]
  \centering
  \includegraphics[width=0.8\linewidth]{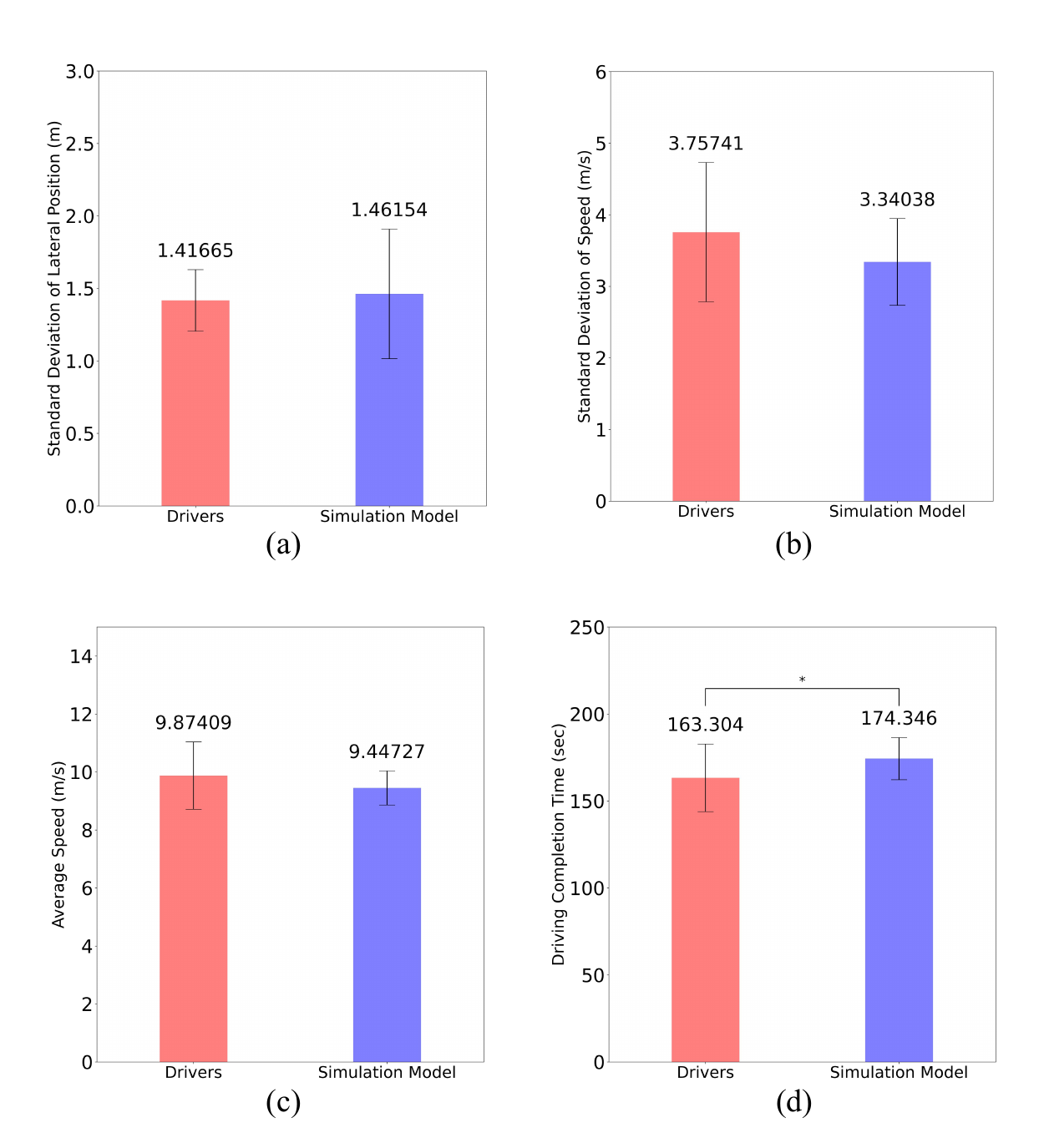}
    \caption{Comparison of average performance of drivers and our simulation model. *: $p < 0.05$, **: $p < 0.005$, ***: $p < 0.0005$}
  \label{fig:sim_eval}
\end{figure*}

Figures \ref{fig:sim_y_x} and \ref{fig:sim_eval} illustrate comparison results of the driving performance of the drivers and the simulation model, respectively. To establish a relationship between the driving performance of drivers and the simulation model, the experiment terrain was segmented into nine sections (See Figure \ref{fig:sim_segmentation}). We examined whether there was a correlation between the average driving performance of drivers and the average driving performance of the simulation model for each of the nine sections. Thus, we conducted a linear regression on the driving performance data for the nine sections and measured the Pearson correlation coefficient ($r$). Figure \ref{fig:sim_y_x} shows the relationship between drivers and the simulation model. Figure \ref{fig:sim_y_x}.a shows that our simulation model is correlated with the drivers in terms of SDLP ($r = 0.84$). Figure \ref{fig:sim_y_x}.b shows that our simulation model is also correlated with the drivers in terms of SDS ($r = 0.81$). Figure \ref{fig:sim_y_x}.c and Figure \ref{fig:sim_y_x}.d demonstrate that our simulation model correlates with drivers in terms of AvgSpeed and DCT ($r = 0.84$ and $r = 0.92$). Based on the results, our simulation model adequately reproduces the teleoperated driving of drivers (Pearson correlation coefficient $r >= 0.8$ for all metrics). 

Figure \ref{fig:sim_eval} illustrates the mean and standard deviation of drivers' SDLP, SDS, AvgSpeed, and DCT, along with the simulation model. We conducted a Welch's T-Test to determine whether a statistically significant difference existed. Figure \ref{fig:sim_eval}.a, Figure \ref{fig:sim_eval}.b, and Figure \ref{fig:sim_eval}.c show that there was no significant difference between the drivers and the simulation model in SDLP ($p=0.64$), SDS ($p=0.06$), and AvgSpeed ($p=0.10$). In contrast, Figure \ref{fig:sim_eval}.d demonstrates that DCT of drivers was significantly less than DCT of the simulation model ($p=0.02$). Although the simulation model performed worse than drivers in DCT, the difference in DCT was 11 seconds. Therefore, we are able to argue that our model adequately simulates drivers' teleoperated driving.

\section{Conclusion}
In this paper, we proposed a CVAE-based teleoperated driving model for unstructured canyon terrains. We implemented a teleoperated driving simulator for unstructured environments, enabling us to mimic teleoperated driving scenarios and collect driving data from drivers. Then, we designed a teleoperated driving model based on a CVAE-based architecture. From time-series data, the forward simulation model and the inverse simulation model generates a perception vector and a control vector at the current time step. The evaluation results demonstrated that our simulation model could reproduce drivers' teleoperated driving. This study could be expanded in the future to include a variety of off-road environments with low-friction roads, craters, and obstacles. If these studies mature, we can anticipate that our proposed simulation model, which mimics human behavior despite using a limited number of user studies, will bring about in a novel approach for dataset generation and user study.

\bibliographystyle{splncs04}
\bibliography{mybibliography}

\begin{thebibliography}{10}
\providecommand{\url}[1]{\texttt{#1}}
\providecommand{\urlprefix}{URL }
\providecommand{\doi}[1]{https://doi.org/#1}

\bibitem{cho2023goondae}
Cho, Y., Yun, H., Lee, J., Ha, A., Yun, J.: Goondae: Denoising-based driver
  assistance for off-road teleoperation. IEEE Robotics and Automation Letters
  \textbf{8}(4),  2405--2412 (2023). \doi{10.1109/LRA.2023.3250008}

\bibitem{choe2015obstacle}
Choe, T.S., Park, J.B., Joo, S.H., Park, Y.W.: Obstacle detection for unmanned
  ground vehicle on uneven and dusty environment. In: Signal Processing,
  Sensor/Information Fusion, and Target Recognition XXIV. vol.~9474, pp.
  433--440. SPIE (2015)

\bibitem{knappe2007driving}
Knappe, G., Keinath, A., Bengler, K., Meinecke, C.: Driving simulator as an
  evaluation tool--assessment of the influence of field of view and secondary
  tasks on lane keeping and steering performance. In: 20th International
  Technical Conference on the Enhanced Safety of Vehicles (ESV) National
  Highway Traffic Safety Administration. No. 07-0262 (2007)

\bibitem{knapper2015comparing}
Knapper, A., Christoph, M., Hagenzieker, M., Brookhuis, K.: Comparing a driving
  simulator to the real road regarding distracted driving speed. European
  journal of transport and infrastructure research  \textbf{15}(2) (2015)

\bibitem{li2022modeling}
Li, C., Tang, Y., Zheng, Y., Jayakumar, P., Ersal, T.: Modeling human steering
  behavior in teleoperation of unmanned ground vehicles with varying speed.
  Human factors  \textbf{64}(3),  589--600 (2022)

\bibitem{li20203d}
Li, N., Yu, X., Liu, X., Lu, C., Liang, Z., Su, B.: 3d-lidar based negative
  obstacle detection in unstructured environment. In: Proceedings of the 2020
  4th International Conference on Vision, Image and Signal Processing. pp.~1--6
  (2020)

\bibitem{moon2023amortized}
Moon, H.S., Oulasvirta, A., Lee, B.: Amortized inference with user simulations.
  In: Proceedings of the 2023 CHI Conference on Human Factors in Computing
  Systems. pp. 1--20 (2023)

\bibitem{murray2022simulation}
Murray-Smith, R., Oulasvirta, A., Howes, A., M{\"u}ller, J., Ikkala, A.,
  Bachinski, M., Fleig, A., Fischer, F., Klar, M.: What simulation can do for
  hci research. Interactions  \textbf{29}(6),  48--53 (2022)

\bibitem{murray2021forward}
Murray-Smith, R., Williamson, J.H., Ramsay, A., Tonolini, F., Rogers, S.,
  Loriette, A.: Forward and inverse models in hci: Physical simulation and deep
  learning for inferring 3d finger pose. arXiv preprint arXiv:2109.03366
  (2021)

\bibitem{naranjo2007cooperative}
Naranjo, J.E., Gonz{\'a}lez, C., Garc{\'\i}a, R., De~Pedro, T.: Cooperative
  throttle and brake fuzzy control for acc $+ $ stop\&go maneuvers. IEEE
  Transactions on Vehicular Technology  \textbf{56}(4),  1623--1630 (2007)

\bibitem{neumeier2019teleoperation}
Neumeier, S., Wintersberger, P., Frison, A.K., Becher, A., Facchi, C., Riener,
  A.: Teleoperation: The holy grail to solve problems of automated driving?
  sure, but latency matters. In: Proceedings of the 11th International
  Conference on Automotive User Interfaces and Interactive Vehicular
  Applications. pp. 186--197 (2019)

\bibitem{opiyo2021review}
Opiyo, S., Zhou, J., Mwangi, E., Kai, W., Sunusi, I.: A review on teleoperation
  of mobile ground robots: Architecture and situation awareness. International
  Journal of Control, Automation and Systems  \textbf{19}(3),  1384--1407
  (2021)

\bibitem{park2021design}
Park, K., Youn, E., Kim, S., Kim, Y., Lee, G.: Design of acceleration feedback
  of ugv using a stewart platform. In: 2021 21st International Conference on
  Control, Automation and Systems (ICCAS). pp. 79--82. IEEE (2021)

\bibitem{schnelle2016driver}
Schnelle, S., Wang, J., Su, H., Jagacinski, R.: A driver steering model with
  personalized desired path generation. IEEE Transactions on Systems, Man, and
  Cybernetics: Systems  \textbf{47}(1),  111--120 (2016)

\bibitem{sohn2015learning}
Sohn, K., Lee, H., Yan, X.: Learning structured output representation using
  deep conditional generative models. Advances in neural information processing
  systems  \textbf{28} (2015)

\bibitem{storms2017shared}
Storms, J., Chen, K., Tilbury, D.: A shared control method for obstacle
  avoidance with mobile robots and its interaction with communication delay.
  The International Journal of Robotics Research  \textbf{36}(5-7),  820--839
  (2017)

\bibitem{wang2020driver}
Wang, Z., Liao, X., Wang, C., Oswald, D., Wu, G., Boriboonsomsin, K., Barth,
  M.J., Han, K., Kim, B., Tiwari, P.: Driver behavior modeling using game
  engine and real vehicle: A learning-based approach. IEEE Transactions on
  Intelligent Vehicles  \textbf{5}(4),  738--749 (2020)

\end{thebibliography}

\end{document}